\documentclass[10pt,twocolumn,letterpaper]{article}

\usepackage{iccv}
\usepackage{times}
\usepackage{epsfig}
\usepackage{graphicx}
\usepackage{amsmath}
\usepackage{amssymb}
\usepackage{graphicx}
\usepackage{booktabs}
\usepackage{algorithmic}
\usepackage[ruled,vlined]{algorithm2e}
\usepackage{multirow}

\usepackage[breaklinks=true,bookmarks=false]{hyperref}
\usepackage{tikz}

\newcommand{\beginsupplement}{%
        \setcounter{table}{0}
        \renewcommand{\thetable}{S\arabic{table}}%
        \setcounter{figure}{0}
        \renewcommand{\thefigure}{S\arabic{figure}}%
     }

\newcommand\copyrighttextfinal{%
    
    \scriptsize\copyright\ 2022 IEEE. Personal use of this material is permitted. Permission from IEEE must be obtained for all other uses, in any current or future media, including reprinting/republishing this material for advertising or promotional purposes, creating new collective works, for resale or redistribution to servers or lists, or reuse of any copyrighted component of this work in other works.}%
\newcommand\copyrightnotice{%
    
    \begin{tikzpicture}[remember picture,overlay]%
        
        \node[anchor=south,yshift=10pt] at (current page.south) {{\parbox{\dimexpr\textwidth-\fboxsep-\fboxrule\relax}{\copyrighttextfinal}}};%
    \end{tikzpicture}%
    
}

\iccvfinalcopy 

\def\iccvPaperID{6}

\ificcvfinal\fi

\begin{document}

\title{SelectNAdapt: Support Set Selection for Few-Shot Domain Adaptation}

\author{Youssef Dawoud $^{1}$ \and Gustavo Carneiro $^{2}$ \and Vasileios Belagiannis $^{1}$\\
\\
$1$ Friedrich-Alexander-Universität Erlangen-Nürnberg, Erlangen, Germany\\
$2$ Centre for Vision, Speech and Signal Processing, University of Surrey, United Kingdom\\
 \\
{\tt\small youssef.dawoud@fau.de}}

\maketitle

\maketitle
\ificcvfinal\thispagestyle{empty}\fi

\begin{abstract}
Generalisation of deep neural networks becomes vulnerable when distribution shifts are encountered between train (source) and test (target) domain data. Few-shot domain adaptation mitigates this issue by adapting deep neural networks pre-trained on the source domain to the target domain using a randomly selected and annotated support set from the target domain. This paper argues that randomly selecting the support set can be further improved for effectively adapting the pre-trained source models to the target domain. Alternatively, we propose \textit{SelectNAdapt}, an algorithm to curate the selection of the target domain samples, which are then annotated and included in the support set. In particular, for the $K$-shot adaptation problem, we first leverage self-supervision to learn features of the target domain data. Then, we propose a per-class clustering scheme of the learned target domain features and select $K$ representative target samples using a distance-based scoring function. Finally, we bring our selection setup towards a practical ground by relying on pseudo-labels for clustering semantically similar target domain samples. Our experiments show promising results on three few-shot domain adaptation benchmarks for image recognition compared to related approaches and the standard random selection. Our code is available at \url{https://github.com/Yussef93/SelectNAdaptICCVW}.
\end{abstract}

\copyrightnotice
\section{Introduction}

Domain shifts between source and target domain data are considered harmful to the generalisation performance of deep neural networks (DNNs). The adaptation of DNNs to the target domain is, thus, essential to preserve their performance on the task in place. Among the family of adaptation methods, few-shot adaptation is a well-known approach that adapts DNNs to the target domain using a few annotated target domain samples. However, few-shot adaptation relies on the random selection of target domain samples to be annotated, which is likely a sub-optimal sample selection procedure.

\begin{figure}[ht]
    \centering
    \includegraphics[width=0.45\textwidth]{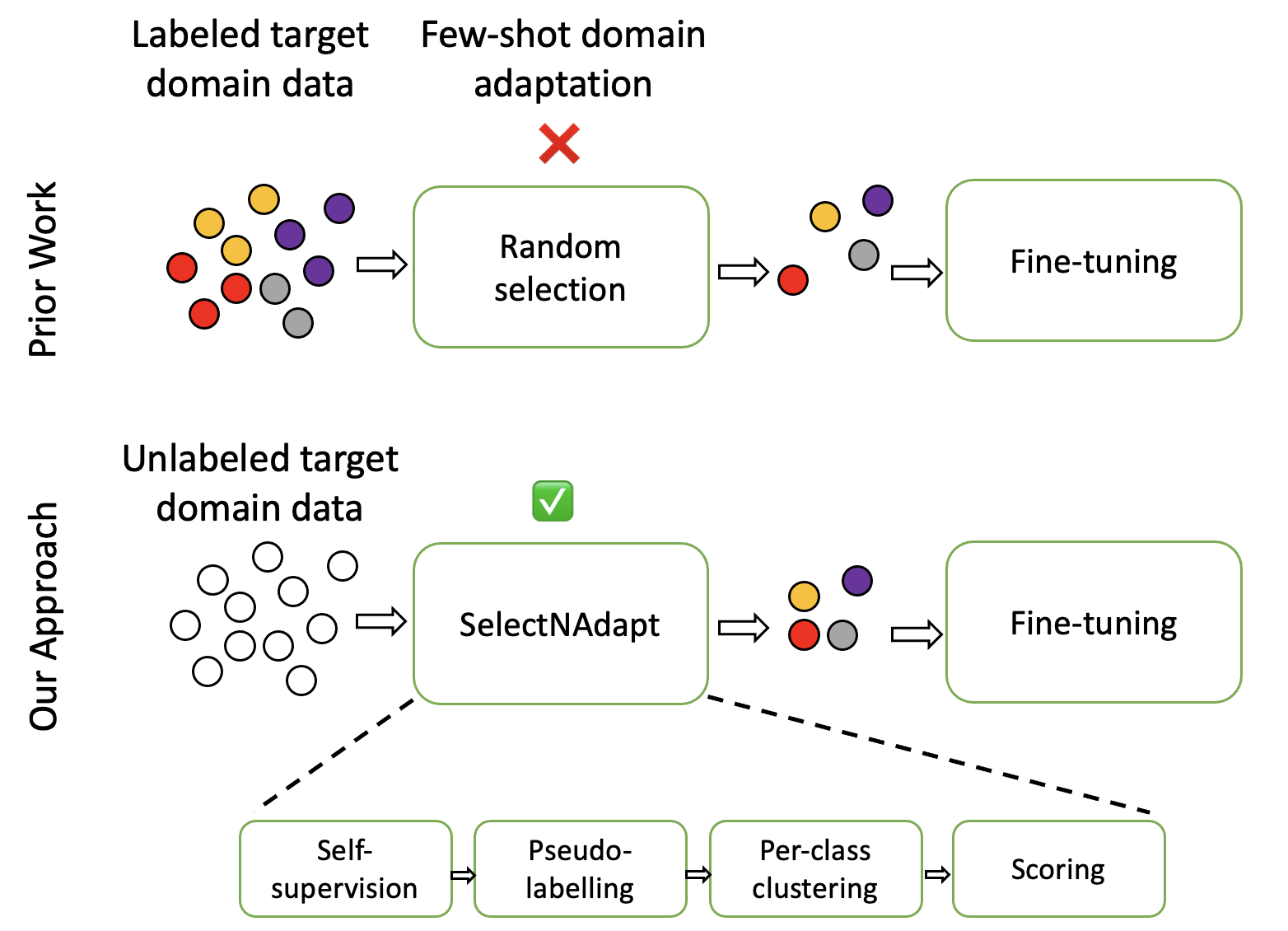}
    \caption{Few-shot domain adaptation is a powerful technique that should be exploited carefully. We propose a more effective support set selection for few-shot domain adaptation by replacing a random selection strategy by an algorithm to select representative target domain samples in an unsupervised way. Our pipeline leverages self-supervision, pseudo-labelling, clustering and selection via a distance metric score.  }
    \label{fig:teaser}
\end{figure}

Adaptation of DNNs can be carried out with varying settings regarding source and target domain data availability. Vanilla (unsupervised) domain adaptation assumes adaptation of DNNs using jointly source and unlabelled target domain data \cite{ganin2015unsupervised}. On the other hand, recent studies argue that access to source domain data at test time is often impractical due to several reasons, including privacy and computation efficiency \cite{wang2020tent}. As a result, test-time domain adaptation emerged as a more interesting alternative setting that disregards source domain data at test-time and assumes only access to the pre-trained source model and the target domain. Well-known test-time adaptation strategies perform on-the-fly adaptation by updating the batch-normalisation (BN) statistics of the source models using unsupervised losses, e.g., entropy minimisation \cite{wang2020tent}. Nevertheless, it has been shown by \cite{ijcai2022p232} that proper adaptation of the BN statistics can not be achieved without supervision from the target domain, as it can not be guaranteed that unsupervised adaptation can correct the domain shifts. Furthermore, test-time adaptation approaches require large mini-batches from the target domain for a good approximation of the BN statistics. Therefore, supervision from the target domain is necessary and can be provided in the form of a small number of randomly selected and annotated target domain samples known as the support set. This process is referred to as source-free few-shot domain adaptation. 

Similarly to few-shot adaptation, few-shot classification tasks assume a class-balanced support set. However, this would require access to the ground-truth of the target domain to select a set of samples per class which is also impractical in real-world situations. Indeed few-shot adaptation has brought a significant improvement compared to the state-of-the-art unsupervised test-time adaptation approaches, yet, it remains unclear whether adaptation of the source model using randomly selected samples from the target domain is sufficient for a good performance on the target domain. Optimising for data sample selection has been thoroughly studied in active learning where data samples are selected and annotated sequentially using unsupervised losses like Shannon's entropy \cite{shannon1948mathematical} or MC-dropout \cite{gal2016dropout}. Nevertheless, it has not been addressed before for few-shot adaptation, where a \textit{support} set is selected in one step only to adapt a pre-trained source model.

In this paper, we focus on few-shot adaptation for image recognition. We empirically argue that proper adaptation of the pre-trained source model requires selecting \textit{representative} target domain data to be included in the support set. Therefore, we propose a simple yet effective selection approach that boosts the few-shot adaptation performance by improving the selection of target samples to be included in the support set. In particular, we propose to perform per-class clustering of the target domain features where the number of clusters is equivalent to the number of $K$-shot adaptation task at hand. The target samples with features close to the cluster centres are included in the support set. However, using the source backbone for extracting target domain's features may negatively impact the clustering and selection process due to the domain shift between source and target. Thus, we seek to narrow this gap by training the source backbone with self-supervision from the target domain. Unlike the prior work \cite{ijcai2022p232}, we rely next on pseudo-labels for determining the target samples of the same pseudo-class to avoid using the target domain ground-truth. Finally, we rely on the Euclidean distance as our selection score to determine the distance of target samples' features to their corresponding cluster centres. A percentage of samples with the smallest distance to the cluster centres are annotated and included in the support set i.e. we use the real labels of support set samples at the adaptation stage.

To the best of our knowledge, we are the first to propose a mechanism to select a support set for few-shot domain adaptation. In summary, our contributions are as follows:
\begin{itemize}

    \item Our algorithm overcomes the target domain shift using state-of-the-art self-supervised tasks to learn target-specific features which aid in a robust support set selection. The learned target features are utilised to select representative samples by per-class clustering of target data features.

     \item We rely on pseudo-labels for the target domain data to select support set samples, in particular, $K$-shots per class. Hence, we ought to perform the selection in a more practical and realistic setup without the need to access the target ground-truth, unlike, prior work.

     \item In our experiments, including several domain adaptation benchmarks, we deliver major improvements compared to random support set selection. Additionally, we show promising results when comparing with other selection approaches, namely, entropy and MC-dropout and few-shot transfer-learning baselines.
\end{itemize}

\section{Related Work}
\label{sec:rw}
\subsection{Few-Shot Domain Adaptation}
Early domain adaptation approaches assumed the availability of source data for jointly adapting a pre-trained deep neural network to a new unlabelled target data \cite{ganin2015unsupervised, pinheiro2018unsupervised}. Under this assumption, several unsupervised-domain adaptation tasks have been developed over the course of the years, including unsupervised-domain classification and segmentation \cite{Saito_2018_CVPR,long2017deep} for natural and medical datasets \cite{guan2021domain, gholami2018novel}. Recently, test-time domain adaptation restricted access to the source data and only allowed access to the pre-trained source model along with the target domain data \cite{wang2022continual,wang2021target,Yang_2021_ICCV}. Test-time adaptation approaches adapt the BN statistics of the source model using unlabelled data from the target domain. For instance, Tent and test-time BN adaptation \cite{nado2020evaluating} rely on unsupervised loss functions like entropy minimisation to update BN parameters using mini-batches from the target domain. Nevertheless, it has been shown in \cite{ijcai2022p232} that proper adaptation of the BN parameters in neural networks requires supervision from the target domain using a randomly selected support set comprising few-annotated samples. In this paper, we argue that random support selection remains ineffective for good approximation of BN statistics. Therefore, we present a method for optimising the selection of support set that further enhances the BN approximation and in return the overall few-shot adaptation performance. Additionally, we pose the selection problem as an unsupervised selection where we rely on pseudo-labels in the per-class clustering stage. In our experiments, we demonstrate state-of-the-art results using our selection mechanism. 

\subsection{Self-Supervised Learning for Domain Adaptation}
Self-supervision is a widely used technique for learning useful representations that enhances the performance on downstream tasks \cite{deshpande2015learning, carlucci2019domain, zhang2019aet, gidaris2018unsupervised}. Over the past few years, self-supervision tasks have been introduced in unsupervised-domain adaptation approaches \cite{Pan_2020_CVPR} where the target domain in conjunction with the supervision from the source domain is utilised to reinforce the representations of the shared backbone network. Accordingly, several self-supervised tasks have been put to practice in unsupervised-domain adaptation and have demonstrated promising results \cite{Kang_2019_CVPR,saito2020universal}. Similarly, contrastive learning has been exploited in test-time adaptation \cite{chen2022contrastive} jointly with pseudo-labels to learn classification on the target domain. Also, in our work, we utilise self-supervised learning. In particular, we train the backbone of the source network using self-supervision defined over the target domain data for reducing the gap between the data features of source and target domains. In return, the learned target features of each class are clustered based on the $K$-shot problem at hand. We show that relying on self-supervision delivers significantly better results than using features of the source backbone.

\section{Method}

\begin{figure*}[ht]
    \centering
    \includegraphics[width=0.8\textwidth]{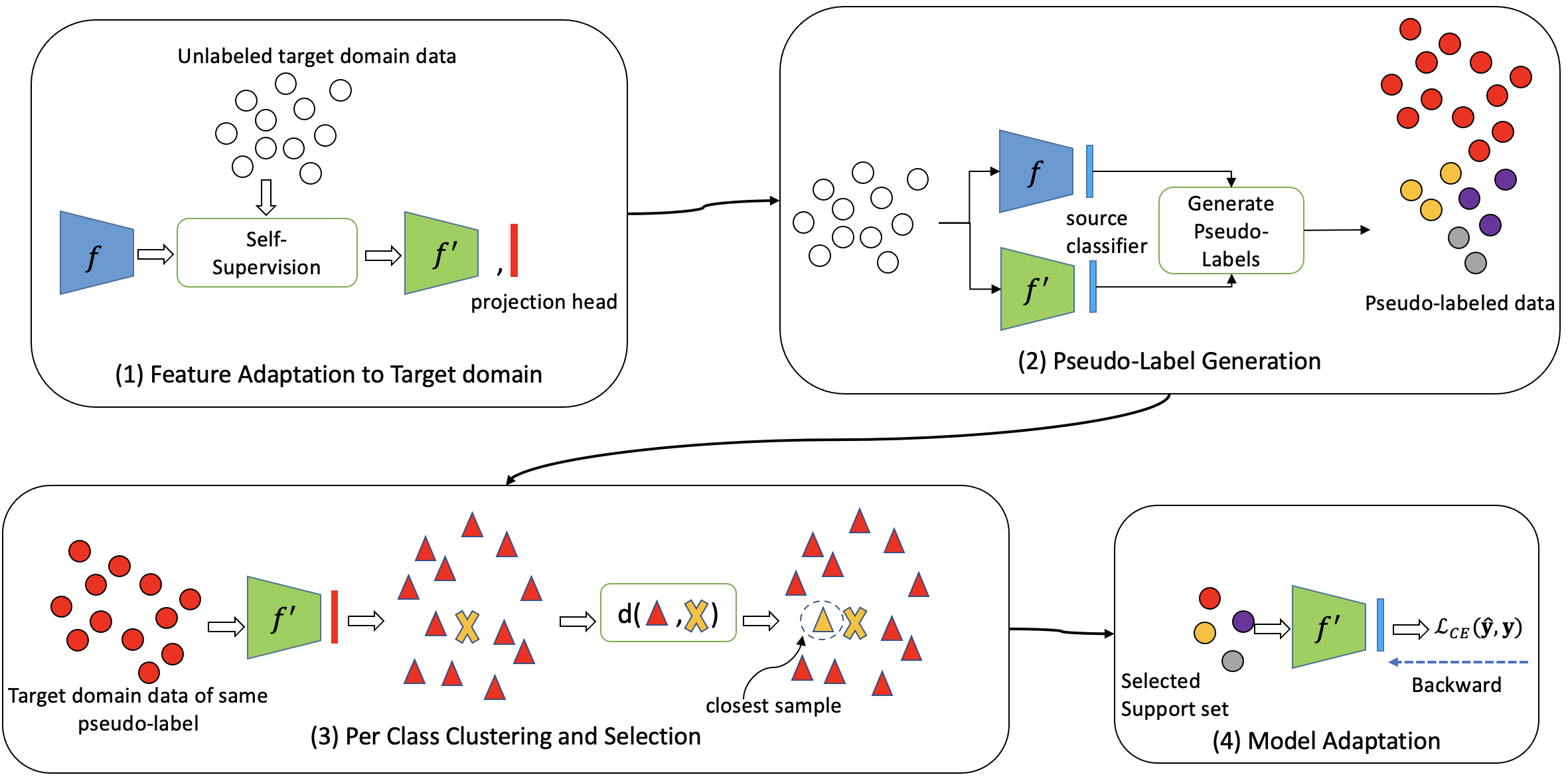}
    \caption{The complete pipeline of our \textit{SelectNAdapt} algorithm at $K=1$-shot. First, we utilise self-supervision for adapting the source features to the target domain data. Then, we generate pseudo-labels with the source classifier for the target domain using the features from the backbone of the pre-trained source model and the backbone trained using BYOL task. Next, we do per-class clustering and calculate the cluster centre (represented by "X") using $K$-means algorithm where the support set samples are selected using the Euclidean distance as the scoring metric. Finally the model is adapted using selected the support set.  }
    \label{fig:my_label}
\end{figure*}

In this section, we start by defining the problem of support set selection. Then, we present our unsupervised support set selection approach for adapting a pre-trained source model to the target domain. 

\begin{algorithm}[ht]
\caption{Unsupervised Support Set Selection}\label{algo}
  \begin{algorithmic}[1]
  \STATE Input: Source model $h_S$ trained using $\mathcal{D}_S$, \\ unlabelled target domain data $\mathcal{D}_T$ , and annotation budget $|\Tilde{\mathcal{D}}_T| = KN$.
 
  \STATE Adapt $f$ with self-supervision using  $\mathcal{D}_T$ and keep $f^\prime , q $. 

\STATE Get pseudo-labels of $\mathbf{X} \in \mathcal{D}_T$ (\ref{psl}).
  \FOR{$c$ = $1,2, \dots, C$}
    \STATE $\mathbf{x}^c = \{\} $
  \FOR{$m$ = $1,2, \dots, |\mathcal{D}_T|$} 
         \IF{$\mathbf{\hat{y}}_m$ = $c$} 
            \STATE $\mathbf{x}^c \cup \{\mathbf{x}_m\}$
            \ENDIF
  \ENDFOR
   
    \STATE Get features $\mathbf{z}^c$ from (\ref{feat}).
    \STATE Cluster $\mathbf{z}^c$ using $K$-means algorithm  i.e. \\ calculate cluster centres $\mu^c = [\mu^{c}_0,\dots,\mu^{c}_{K-1}]$. 
    \FOR{$i$= 1,2,\dots,$K$}
    \STATE Calculate $d(\mathbf{z}^{c,i}, \mu^{c}_i)$ from (\ref{dist}).
    
    \ENDFOR
     \STATE Select $\mathcal{\Tilde{D}}_T$ from (\ref{score}) using $d(\mathbf{z}^c,\mathbf{\mu}^c)$. 

 \ENDFOR
 \STATE Update BN parameters of $f^\prime$ using LCCS and $\mathcal{\Tilde{D}}_T$ from (\ref{adapt}).
 \STATE Output: Evaluate $h_T$ on $\mathcal{\hat{D}}_T $.
 \end{algorithmic}
 \end{algorithm}

\label{sec:method}
\subsection{Problem Definition}
Let $h_{S} = g \circ f$ be a deep neural network trained on the source domain $\mathcal{D}_S$, where $f$ denote the backbone network of the source model that maps an input image to a latent code (feature representation) $f: \mathcal{X} \rightarrow \mathcal{Z}, \mathcal{Z} \subset \mathbb{R}^D$ and $g$ is the task head network that maps features extracted by $f$ to the output space of the learning task at hand $g: \mathcal{Z} \rightarrow\mathcal{Y}$. In this work, we focus on image recognition, hence, $g$ is a classification head that maps the features to the label space $\mathcal{Y} \subset [0,1]^C$, where $C$ is the total number of classes. Note that both source and target images share the same label space. Given access only to $h_S$ and the target domain $\mathcal{D}_T$ at adaptation time, our objective is to seek for few annotated target samples, namely, the support set $\mathcal{\Tilde{D}}_T \subset \mathcal{D}_T$, to adapt $h_S$ to the shifted target domain $\mathcal{D}_T$. In general, a few-shot classification task is framed as $C$-way, $K$-shot task which is referred to as the support set, where $C$ is the number of semantic classes and $K$ is the number of samples per class, as a rule of thumb $K \leq 10$. Instead of randomly selecting the target samples, we present a support set selection algorithm that improves the adaptation performance compared to random selection. The size of $\mathcal{\Tilde{D}}_T$ is set to $KC$. 

To reach our goal, we learn features of the target domain data by training the backbone network of the source model on state-of-the-art self-supervised tasks, namely, contrastive learning task \cite{grill2020bootstrap, caron2020unsupervised}. Next, we get pseudo-labels of the target domain data using the features of both the source backbone and the backbone trained using self-supervision. Afterwards, we perform per-class clustering using the features learned from the self-supervised task and select the target samples according to our scoring function, which is the minimum Euclidean distance to the cluster centre. In the end, we adapt the model to the target domain using our selected support set and evaluate the adapted model on the target test set. We present our algorithm in detail below. 

\subsection{SelectNAdapt Algorithm}
As previously stated, we assume access to the pre-trained source model $h_{S}$ and the target domain $\mathcal{D}_T$ of size $M$ containing unlabelled images denoted by $\mathbf{X}$. We summarise the steps of our approach in algorithm \ref{algo}. Moreover, a visual explination is provided in Fig. \ref{fig:my_label}.

\subsubsection{Feature Adaptation} We argue that source-extracted features of target domain data may prohibit an effective selection of support samples in the per-class clustering step due to the shifted target domain. To alleviate this issue, we train the source backbone $f \rightarrow f^\prime$ using self-supervision to learn target-related features, which provisions a better feature clustering compared to using the source backbone and, thus, better support set selection. Contrastive learning \cite{chen2020simple,caron2020unsupervised,grill2020bootstrap} have gained a wide reputation over the past years for their ability to learn useful representation. In this context, the objective of the learning task is to train the backbone of a DNN with a projection head attached to it to learn an embedding space that pulls similar data pairs together while pushing dissimilar ones apart. Afterwards, the backbone is fine-tuned on a particular downstream task. Accordingly, we use contrastive learning tasks to train the source backbone using the target domain data $\mathcal{D}_T$. After training we keep the trained backbone $f^\prime$ and the projection head $q$, which projects the features extracted by $f^\prime$ onto a $d$-dimensional feature space where $d < D$.

\subsubsection{Pseudo-labels generation} A few-shot classification task is defined as $C$-way, $K$-shot i.e. the support set should contain $K$-shots for every class $c \in C$. To construct the support set from unlabelled target data, we first generate pseudo-labels for $\mathbf{X}$ by using the features of $f$ and $f^\prime$ along with the source classifier. To obtain the pseudo-labels, we follow an ensemble prediction model \cite{sagi2018ensemble} where we average the output probability distributions of the source classifier $g$ using the features of $f$ and $f^\prime$. The assigned pseudo-label is based on the maximum output probability over the distribution of classes which we define as follows:
\begin{equation}
    \mathbf{\hat{Y}} = \arg \max_{1,2,\dots,C} \frac{\sigma(g(f(\mathbf{X}))) + \sigma(g(f^{\prime}(\mathbf{X})))}{2},
    \label{psl}
\end{equation}
where $\mathbf{\hat{Y}} = [0,1]^{M \times C}$ is a matrix that holds one-hot encoding vectors of length $C$ for all the target sample and $\sigma$ is a softmax activation function \cite{bishop2006pattern}. Afterwards, we group the target samples according to their pseudo-labels into $C$ categories, i.e. $\mathbf{X} = [\mathbf{x}^0,\dots,\mathbf{x}^{C-1}]$, with $\mathbf{x}^c = [\mathbf{x}_{1}^c,\dots,\mathbf{x}_{|\mathbf{x}^c|}^c ], c \in \mathbf{\hat{Y}}$ and extract their features $\mathbf{z}^{c}$ using $f^\prime$:

\begin{equation}
    \mathbf{z}^c = q(f^{\prime}(\mathbf{x}^c)), \quad \forall \mathbf{x}^c \in \mathbf{X},
    \label{feat}
\end{equation}
where $\mathbf{x}^c$ is the set of all target domain samples with pseudo-label $c$, $\mathbf{z}^c$ is a matrix containing $d$-dimensional feature vectors of $\mathbf{x}^c$, and $q$ is the projection network retained from the self-supervised learning task. It is noteworthy that adding a projection head in contrastive learning frameworks empirically performs better than relying on the raw features of the backbone network \cite{chen2020simple}. Therefore, we leverage the output features of the projection network in the clustering step.

\subsubsection{Per-Class Clustering and Selection} We rely on a scoring function to rank and select target data samples per class. To this end, we cluster $\mathbf{z}^c$ into $K$-clusters. Note that $K$ is the number of shots per class $c$. We calculate the cluster centres $\mu^c = [\mu^{c}_1,\dots,\mu^{c}_{K}]$ based on the $K$-means algorithm \cite{pedregosa2011scikit} and score the target samples according to the distance of their features to the cluster centres. We make use of the Euclidean distance as our scoring function that measures the distance between the features assigned to cluster $i$ and their cluster centre $\mu^{c}_i$, where $i \in [1,\dots,K]$, we define the distance metric:
\begin{equation}
d(\mathbf{z}^{c,i}, \mu^{c}_i) = ||\mathbf{z}^{c,i} - \mu^{c}_i||_2.
\label{dist}
\end{equation}

The features of the target samples with the minimum distance to their corresponding cluster centre are included in the support set $\mathcal{\Tilde{D}_T}$, hence, our selection metric becomes:
\begin{equation}
\begin{split}
    \mathcal{\Tilde{D}}_T = \arg \min_{\mathcal{D}_T} \sum_{c=0}^{C-1} \sum_{i=0}^{K-1} \sum_{z^c \in \mu_{i}^c} d(\mathbf{z}^{c,i}, \mu^{c}_i) \\ s.t. |\mathcal{\Tilde{D}}_T| = KN.
\end{split}
    \label{score}
\end{equation}

Next, the selected samples to be included in the support set are associated with their true labels i.e. we do not use their pseudo-labels at adaptation time, hence, $\mathcal{\Tilde{D}}_T = \{(\mathbf{x}_j,\mathbf{y}_j)\}_{j=1}^{KN}$, where $\mathbf{y}_j$ is the true label of $\mathbf{x}_j$. 

\subsubsection{Model adaptation}
Eventually, we use $\mathcal{\Tilde{D}}_T$ to update the BN parameters of the pre-trained source model where its backbone network is replaced with $f^\prime$. We follow the approach of linear combination coefficients for batch normalisation statistics (LCCS) \cite{ijcai2022p232} to update the BN parameters using the cross-entropy loss \cite{bishop2006pattern} $\mathcal{L}_{CE}(g,f^\prime,\mathcal{\Tilde{D}}_T)$ , hence the update BN parameters are optimised as:

\begin{equation}
    \theta^* = \arg \min_{\theta_{f^\prime}} \mathcal{L}_{CE}(g,\theta_{f^\prime},\mathcal{\Tilde{D}}_T),
    \label{adapt}
\end{equation}

where $\theta_{f^\prime}$ denote the BN parameters of the backbone trained on BYOL and $\theta^*$ are the updated BN parameters. Finally, we evaluate the updated model ($h_T$) on the target test set $\mathcal{\hat{D}}_T =\mathcal{D}_T \setminus \mathcal{\Tilde{D}}_T$.

\section{Experiments}

\label{sec:exp}
\begin{table}[ht]
\centering
\begin{tabular}{|l|c|l|l|l|}
\hline
\multicolumn{1}{|c|}{Dataset} & PACS     & VisDA                               & Office-31                                                \\ \hline
Backbone                   & ResNet-18 & \multicolumn{1}{c|}{ResNet-101} & \multicolumn{1}{c|}{ResNet-50}  \\ \hline
\end{tabular}
\caption{Backbone architectures used in our experiments.}
\label{tab:arch}
\end{table}
\begin{table*}[ht]
\centering
\begin{tabular}{|l|ccccccc|}
\hline
\multicolumn{1}{|c|}{\multirow{2}{*}{Support Set Selection Method}} &
  \multicolumn{7}{c|}{Office-31} \\ \cline{2-8} 
   &
  \multicolumn{1}{l|}{A $\rightarrow$ W} &
  \multicolumn{1}{l|}{A $\rightarrow$D} &
  \multicolumn{1}{l|}{W $\rightarrow$A} &
  \multicolumn{1}{l|}{W $\rightarrow$D} &
  \multicolumn{1}{l|}{D$\rightarrow$ W} &
  \multicolumn{1}{l|}{D $\rightarrow$A} &
  \multicolumn{1}{l|}{Average}\\ \hline
Random \cite{ijcai2022p232}&
  \multicolumn{1}{c|}{92.8} &
  \multicolumn{1}{c|}{91.8} &
  \multicolumn{1}{c|}{75.1} &
  \multicolumn{1}{c|}{99.9} &
  \multicolumn{1}{c|}{98.5} &
  \multicolumn{1}{c|}{75.4} &
  88.9 \\ \hline
Entropy &
  \multicolumn{1}{c|}{88.4} &
  \multicolumn{1}{c|}{87.7} &
  \multicolumn{1}{c|}{72.3} &
  \multicolumn{1}{c|}{97.7} &
  \multicolumn{1}{c|}{97.3} &
  \multicolumn{1}{c|}{71.9} &
  85.9 \\ \hline
MC-dropout &
  \multicolumn{1}{c|}{87.5} &
  \multicolumn{1}{c|}{85.8} &
  \multicolumn{1}{c|}{73.8} &
  \multicolumn{1}{c|}{100.0} &
  \multicolumn{1}{c|}{98.2} &
  \multicolumn{1}{c|}{71.9} &
  85.9 \\ \hline

\textit{\textbf{Ours}} &
  \multicolumn{1}{c|}{95.0} &
  \multicolumn{1}{c|}{97.4} &
  \multicolumn{1}{c|}{76.4} &
  \multicolumn{1}{c|}{100.0} &
  \multicolumn{1}{c|}{99.7} &
  \multicolumn{1}{c|}{75.9} &
  \textbf{90.7} \\ \hline
\end{tabular}
\caption{A comparison of adaptation test results using random, entropy, MC-dropout and \textit{SelectNAdapt} algorithm (\textbf{\textit{Ours}}) for Office-31 dataset at $K$ = 5-shot.}
\label{tab:office31}
\end{table*}

\begin{table*}[ht]
\centering
\begin{tabular}{|l|ccc|ccc|}
\hline
\multirow{2}{*}{Support Set Selection Methods} & \multicolumn{3}{c|}{PACS}                        & \multicolumn{3}{c|}{VisDA}                       \\ \cline{2-7} 
 & \multicolumn{1}{c|}{1-shot} & \multicolumn{1}{c|}{5-shot} & 10-shot & \multicolumn{1}{c|}{1-shot} & \multicolumn{1}{c|}{5-shot} & 10-shot \\ \hline
Random \cite{ijcai2022p232}                 & \multicolumn{1}{c|}{81.6} & \multicolumn{1}{c|}{86.1} & 87.6 & \multicolumn{1}{c|}{67.8} & \multicolumn{1}{c|}{76.0} & \textbf{79.2} \\ \hline
Entropy                 & \multicolumn{1}{c|}{79.6} & \multicolumn{1}{c|}{86.2} &  86.1& \multicolumn{1}{c|}{61.0} & \multicolumn{1}{c|}{73.8} & 74.2  \\ \hline
MC-dropout              & \multicolumn{1}{c|}{80.3} & \multicolumn{1}{c|}{86.1} & 86.8 & \multicolumn{1}{c|}{68.1} & \multicolumn{1}{c|}{73.1} & 74.1 \\ \hline
\textbf{\textit{Ours}}                    & \multicolumn{1}{c|}{\textbf{84.5}} & \multicolumn{1}{c|}{\textbf{87.9}} & \textbf{87.9} & \multicolumn{1}{c|}{\textbf{73.0}} & \multicolumn{1}{c|}{\textbf{78.0}} & 78.0 \\ \hline
\end{tabular}
\caption{A comparison of averaged few-shot adaptation test results using different selection approaches, namely, random, entropy, MC-dropout and \textit{SelectNAdapt} algorithm (\textbf{\textit{Ours}}) for PACS and VisDA datasets. Note that we average adaptation results over the target domains of PACS.}
\label{tab:pvc}
\end{table*}

\begin{table*}[ht]
\centering
\begin{tabular}{|l|ccc|ccc|}
\hline
\multirow{2}{*}{Few-Shot Adaptation Methods}   & \multicolumn{3}{c|}{PACS}                        & \multicolumn{3}{c|}{VisDA}                       \\ \cline{2-7} 
 &
  \multicolumn{1}{c|}{1-shot} &
  \multicolumn{1}{l|}{5-shot} &
  \multicolumn{1}{l|}{10-shot} &
  \multicolumn{1}{c|}{1-shot} &
  \multicolumn{1}{l|}{5-shot} &
  \multicolumn{1}{l|}{10-shot} \\ \hline
Ada-BN  \cite{li2016revisiting}                 & \multicolumn{1}{c|}{82.9} & \multicolumn{1}{c|}{85.5} &  85.8& \multicolumn{1}{c|}{56.5} & \multicolumn{1}{c|}{60.9} & 61.8 \\ \hline
fine-tune BN  \cite{li2016revisiting}            & \multicolumn{1}{c|}{79.0} & \multicolumn{1}{c|}{84.3 } & 85.4 & \multicolumn{1}{c|}{59.1  } & \multicolumn{1}{c|}{70.9} & 74.9\\ \hline
fine-tune classifier \cite{li2016revisiting}     & \multicolumn{1}{c|}{82.5} & \multicolumn{1}{c|}{83.7} &  83.8 & \multicolumn{1}{c|}{67.6  } & \multicolumn{1}{c|}{69.7} & 77.4 \\ \hline
fine-tune feat. extractor  \cite{li2016revisiting}& \multicolumn{1}{c|}{83.6 } & \multicolumn{1}{c|}{86.0 } & 86.1 & \multicolumn{1}{c|}{67.3  } & \multicolumn{1}{c|}{68.4} & 74.7 \\ \hline
L$^2 $  \cite{xuhong2018explicit}                     & \multicolumn{1}{c|}{84.4 } & \multicolumn{1}{c|}{85.8} &  85.6 & \multicolumn{1}{c|}{66.0 } & \multicolumn{1}{c|}{66.4 } &  69.6\\ \hline
L$^2 $-SP  \cite{xuhong2018explicit}               & \multicolumn{1}{c|}{84.4} & \multicolumn{1}{c|}{85.8} & 85.6 & \multicolumn{1}{c|}{66.0  } & \multicolumn{1}{c|}{66.4} & 69.6  \\ \hline
DELTA  \cite{li2019delta}                   & \multicolumn{1}{c|}{84.4 } & \multicolumn{1}{c|}{85.8} & 85.6 & \multicolumn{1}{c|}{65.9  } & \multicolumn{1}{c|}{66.5} & 70.1 \\ \hline
Late Fusion \cite{latefusion}              & \multicolumn{1}{c|}{83.2  } & \multicolumn{1}{c|}{83.6} & 83.6 & \multicolumn{1}{c|}{67.2  } & \multicolumn{1}{c|}{69.8} & 74.5 \\ \hline
FLUTE   \cite{triantafillou2021learning}                  & \multicolumn{1}{c|}{73.4} & \multicolumn{1}{c|}{85.8} & 88.1 & \multicolumn{1}{c|}{48.3  } & \multicolumn{1}{c|}{67.1} & 65.7  \\ \hline
LCCS  \cite{ijcai2022p232}                  & \multicolumn{1}{c|}{84.4  } & \multicolumn{1}{c|}{87.1} & 88.8 & \multicolumn{1}{c|}{67.8} & \multicolumn{1}{c|}{76.0} & \textbf{79.2}  \\ \hline
\textbf{\textit{Ours}}                     & \multicolumn{1}{c|}{\textbf{88.2}} & \multicolumn{1}{c|}{\textbf{89.3}} & \textbf{89.5} & \multicolumn{1}{c|}{\textbf{73.0}} & \multicolumn{1}{c|}{\textbf{78.0}} &  78.0 \\ \hline
\end{tabular}
\caption{We report test results averaged over PACS target domains, as well as the test results of VisDA for $K =$ 1-, 5-, 10-shots. In this table, we compare against different few-shot transfer learning approaches tailored to the setting of source-free few-shot domain adaptation. Note that the pre-trained source model $f$ has been trained with MixStyle domain generalisation approach \cite{zhou2021domain} on PACS dataset.   }
\label{tab:baseline_compare}
\end{table*}

\subsection{Datasets}
We conduct our experiments using three domain adaptation benchmarks for image recognition. Namely, we use PACS (4 domains with 7 classes) \cite{li2017deeper}, Office-31 (3 domains with 31 classes) \cite{saenko2010adapting}, and VisDA datasets (2 domains with 12 classes) \cite{visda2017}. Each of these datasets incurs domain shifts in the form of different image styles (PACS and VisDA) or images captured with different cameras (Office-31). We adopt the same evaluation metrics used in \cite{ijcai2022p232}, in particular, we adopt accuracy as an evaluation metric for PACS, average-per-class accuracy for Office-31, and average precision for VisDA. The evaluation protocol for PACS and VisDA follows a leave-one-domain-out cross validation \cite{dou2019domain} where one domain is left out as the target domain and the rest is treated as the source domain(s). On the other hand, the evaluation protocol for Office-31 splits the dataset into 6 pairs, each pair contains one source domain and one target domain.

\subsection{Implementation Details}
\paragraph{Source Models} We use different backbone networks for each dataset as shown in Tab.\ref{tab:arch}. Our source models are trained on the source domain(s) using empirical risk minimisation (ERM) \cite{gulrajani2021in}. However, we use the publicly available pre-trained model CSG ResNet-101 \cite{chen2021contrastive} on the source domain of VisDA. As for PACS and Office-31 we reproduce the training on the source domains following the implementations of \cite{zhou2022domain,zhou2021domain}. 

\paragraph{Feature Adaptation} As previously stated, we rely on contrastive learning, namely BYOL, a regressive self-supervised task. BYOL is a well-known self-supervised contrastive learning framework that does not require negative samples and is less sensitive to hyper-parameter changes. In BYOL, two networks with identical backbones, namely, online and target networks interact and learn from each other. In particular, the online network learns to regress the features of the target network under different augmentations of the same image. Hence, it enforces consistent representations. We initialise the backbones of the online and target networks with the parameters of $f$ and learn the BYOL task. For PACS dataset, we train for 100 epochs using a LARS optimiser \cite{you2017scaling} with initial learning rate of 0.2 and cosine annealing scheduler \cite{loshchilov2017sgdr}. As for the remaining datasets, we use an Adam optimiser \cite{kingma2014adam} with learning rate of 0.0001. Moreover, we train for 100 epochs for the target domains of Office-31. However, for VisDA we empirically observed that 10 epochs are sufficient for the training to converge on the BYOL task. For all target domains, we use a mini-batch size of 256. Upon completing the learning task we keep the backbone and the projection head of the online network and use them for extracting features of target domain data.

\paragraph{Support Set Selection} We compare our selection against random, entropy, and MC-dropout selection approaches. For random selection, following \cite{ijcai2022p232}, the target domain data are chosen according to a uniform probability distribution. Note that the support set in \cite{ijcai2022p232} is class-balanced i.e. there is an equal number of support samples for every class in the target domain. As for entropy, we calculate Shannon's entropy using the softmax output of the source model \cite{shannon1948mathematical}. For each class in the target data, we select the top-$K$ samples with the highest entropy loss. Similarly, for MC-dropout we average Shannon's entropy over 10 forward passes using $h_S$ with a dropout layer of probability 0.5 inserted at the final layer of the backbone \cite{gal2016dropout}. This setting has been shown to yield the best result. We also compare our selection approach to few-shot transfer learning baselines, which adapts the pre-trained source model in different ways. 

\paragraph{Model adaptation} The BN parameters are adapted based on LCCS method of \cite{ijcai2022p232} using an Adam optimiser for 10 epochs with 0.001 learning rate with mini-batch size of 32. During adaptation on datasets PACS and VisDA we use a nearest-centroid classifier \cite{yoo2018efficient} for $K \geq 5$ \cite{triantafillou2021learning}, otherwise, we use the pre-trained source classifier. However, for Office-31, the source classifier is fine-tuned after adapting the BN parameters for 200 epochs using the same Adam optimiser settings for adapting the BN parameters. We average the test results over at least 3 different seeds.

\subsection{Support Set Selection Comparison}

 We present our averaged numerical test results for random, entropy, MC-dropout and our selection approach in Tab. \ref{tab:pvc} and \ref{tab:office31} . For PACS datasets, we also average the test results over the target domains for better comparison of different selection approaches. Results per domain could be viewed in the supplementary material. Clearly, our approach mainly dominates all other selection approaches on all the benchmarks which supports our claim that careful selection of support set samples from the target is vital for an effective adaptation performance. Although our approach may result in a class-imbalanced support set due to false pseudo-labels that do not align with the real labels of target samples, we notice that adaptation performance remains robust and still performs better compared to the random selection, which is yet an additional reason that highlights the importance of representative support set samples over a randomly selected and class-balanced support set. Furthermore, we observed in several cases that random selection could have a better performance than entropy and MC-dropout approaches. We attribute this behaviour to the tendency of entropy to select samples that lie close to the decision boundaries of per-class clusters, which result in high prediction uncertainty. These samples are less beneficial for the adaptation performance as they are biased towards a specific region i.e. the boundaries of the decision space. Hence, they are considered poor representative candidates for the adaptation task of BN parameters. On the other hand, our approach that learns target features and then performs per-class clustering to select target domain samples that fall near the cluster centres, i.e. samples that represent each cluster, positively impacts the adaptation performance.

\subsection{Few-Shot Learning Comparison}
In Tab. \ref{tab:baseline_compare}, we report the results of few-shot transfer learning approaches tailored to fit the setting of few-shot adaptation, which neglects the presence of source domain data at adaptation time and adapts the pre-trained source model using a randomly selected support set. These approaches include AdaBN \cite{li2016revisiting} for replacing BN statistics using a randomly selected support set from the target domain. Later on either the BN parameters of the source model or the backbone network or classifier layer are fine-tuned. Moreover, other approaches like L$^2$, L$^2$-SP \cite{xuhong2018explicit}, DELTA \cite{li2019delta} that fine-tunes the entire source model using an additional regularisation term are shown. Finally, we have FLUTE \cite{triantafillou2021learning} that adapts BN parameters using nearest-centroid classifier and Late Fusion \cite{latefusion} which averages classification results using source and target classifier. We employ the source model trained using the MixStyle approach \cite{zhou2021domain} for a fair comparison with the random selection baseline, . We observe that our selection method can still show quite significant results compared to the random selection and the few-shot learning baselines, which highlights the significance of proper support set selection compared to using different adaptation techniques. In the following section, we conduct more ablation studies using the PACS dataset to analyse various components of our approach.

\subsection{Ablation Study}
\begin{table*}[ht]
\centering
\begin{tabular}{|l|c|c|ccc|}
\hline
\multirow{2}{*}{Selection Method} & \multirow{2}{*}{Adapted Model} & \multirow{2}{*}{Bal} & \multicolumn{3}{c|}{PACS}                                             \\ \cline{4-6} 
                        &    &  & \multicolumn{1}{c|}{1-shot} & \multicolumn{1}{c|}{5-shot} & 10-shot               \\ \hline
\multirow{2}{*}{Random \cite{ijcai2022p232}}  & $f$  & \checkmark & \multicolumn{1}{c|}{81.6} & \multicolumn{1}{c|}{86.1} &      87.6              \\ \cline{2-6} 
                        & $f^\prime$ & \checkmark & \multicolumn{1}{c|}{84.3} & \multicolumn{1}{c|}{85.9} & \multicolumn{1}{c|}{87.9}                \\ \hline
\multirow{2}{*}{\textbf{\textit{Ours}} }             & $f^\prime$                             &                  x    & \multicolumn{1}{c|}{84.5} & \multicolumn{1}{c|}{87.9} &     87.9   \\ \cline{2-6} 
                        & $f^\prime$ & \checkmark & \multicolumn{1}{c|}{85.6} & \multicolumn{1}{c|}{88.4} & \multicolumn{1}{c|}{88.6} \\ \hline
\end{tabular}
\caption{A comparison of few-shot adaptation test results for random selection and our approach under different settings of adapting BYOL trained backbone, i.e., $f^\prime$ and the source backbone $f$, in addition to, class-balanced (Bal) support sets.}
\label{tab:ab1}
\end{table*}

\begin{table}[ht]
\centering
\begin{tabular}{|c|cll|}
\hline
\multirow{2}{*}{\begin{tabular}[c]{@{}l@{}}SelectNAdapt\\ \quad   $f$  \quad \quad $f^\prime$    \end{tabular}} &
  \multicolumn{3}{c|}{PACS} \\ \cline{2-4} 
 &
  \multicolumn{1}{c|}{1-shot} &
  \multicolumn{1}{c|}{5-shot} &
  \multicolumn{1}{c|}{10-shot} \\ \hline
 \checkmark \quad \quad x & \multicolumn{1}{c|}{79.6}       & \multicolumn{1}{c|}{85.5} & \multicolumn{1}{c|}{87.6  } \\  \hline
\checkmark \quad \quad \checkmark  & \multicolumn{1}{c|}{84.5}       & \multicolumn{1}{c|}{87.9}       &    \multicolumn{1}{c|}{87.9}               \\ \hline
\end{tabular}
\caption{We compare the few-shot adaptation results of using only the backbone of the pre-trained source model ($f$) for pseudo-label generation, per-class clustering, and model adaptation against using additionally self-supervision i.e. the backbone trained using BYOL ($f^\prime$). }
\label{tab:ab2}
\end{table}

\begin{table}[ht]
\centering
\begin{tabular}{|l|ccc|}
\hline
\multirow{2}{*}{Self-Supervision} & \multicolumn{3}{c|}{PACS}                                           \\ \cline{2-4} 
                        & \multicolumn{1}{c|}{1-shot} & \multicolumn{1}{c|}{5-shot} & 10-shot \\ \hline
BYOL \cite{grill2020bootstrap}                  & \multicolumn{1}{c|}{84.5}       & \multicolumn{1}{c|}{87.9}       &  87.9       \\ \hline
SwAV \cite{caron2020unsupervised}                   & \multicolumn{1}{c|}{86.1}       & \multicolumn{1}{c|}{87.7}       &  88.2       \\ \hline
\end{tabular}
\caption{A comparison of averaged few-shot adaptation test results for training the source backbone with BYOL and SwAV in our support set selection pipeline on the target domains of PACS dataset.}
\label{tab:ab3}
\end{table}

\begin{table}[]
\centering
\begin{tabular}{|c|cll|}
\hline
\multirow{2}{*}{\begin{tabular}[c]{@{}l@{}}Pseudo-labelling\\ \quad \enskip  $f$  \quad \quad $f^\prime$    \end{tabular}} &
  \multicolumn{3}{c|}{PACS} \\ \cline{2-4} 
 &
  \multicolumn{1}{c|}{1-shot} &
  \multicolumn{1}{c|}{5-shot} &
  \multicolumn{1}{c|}{10-shot} \\ \hline
 \checkmark \quad \quad x & \multicolumn{1}{c|}{84.6} & \multicolumn{1}{c|}{87.5} & \multicolumn{1}{c|}{87.7} \\ \hline
x \quad \quad \checkmark  & \multicolumn{1}{c|}{84.3}       & \multicolumn{1}{c|}{87.4}       &    \multicolumn{1}{c|}{87.9}                        \\ \hline
\checkmark \quad \quad \checkmark  & \multicolumn{1}{c|}{84.5} & \multicolumn{1}{c|}{87.9} &     \multicolumn{1}{c|}{87.9}                       \\ \hline
\end{tabular}
\caption{We show a comparison of averaged few-shot adaptation test results for using different combinations of source $f$ and self-supervised (BYOL) backbones $f^\prime$ to generate pseudo-labels. }
\label{tab:ab5}
\end{table}
\paragraph{Class-balanced support set and model adaptation} As previously mentioned, the random selection baseline \cite{ijcai2022p232} ensures a class-balanced support set following the few-shot classification protocol of $C$-way, $K$-shot. To achieve this, prior knowledge of target domain data ground-truth is essential to select $K$-shots randomly from the set of target samples belonging to a particular class $c \in C$. Accordingly, we carry out an experiment assuming a prior knowledge of target domain ground-truth. In this case, we skip the pseudo-labelling step and use directly the self-supervised model $f^\prime$ to extract target data features and perform per-class clustering to find representative target samples to be included in the support set. The averaged results on the PACS datasets are shown in Tab. \ref{tab:ab1} where it is noticeable that a class-balanced support set widens the performance gap between our approach and random selection. Hence, we deduce from this experiment that applying our selection for finding representative target samples is more efficient compared to random selection even in the presence of ground-truth data. Additionally, we analyse the impact of our selection approach against random selection using $f^\prime$ as the backbone to be adapted. We notice the overall performance remains better using our selection approach, especially at $K=1$-shot, implying a more efficient selection mechanism compared to random selection. On the other hand, the performance of random selection also increases relative to using the pre-trained source backbone since the self-supervised pre-trained backbone yields a better initialisation of network parameters due to the learned representation on the target domain. Finally, we observe the performance gap between the random selection and our approach is reduced as $K$ increases due to the availability of sufficient training data.


\paragraph{Source model for pseudo-label generation, per-class clustering and selection} In Tab.\ref{tab:ab2}, we document the results of neglecting the self-supervision step. Specifically, we use the source backbone for generating pseudo-labels, per-class clustering, and selecting support set samples from target domain data, and adapting it using the selected support set. From the results, we notice a significant difference between using the backbone of the source model alone and the backbone trained on the BYOL task. This clearly indicates the impact of self-supervision in the selection process as it bridges the domain shift gap between source and target domains by learning useful target features for a robust selection of the support set. Furthermore, in Tab. \ref{tab:ab5}, we study the effect of using different combinations of source and self-supervised (BYOL) backbones i.e. $f$ and $f^\prime$, to generate pseudo-labels. Combining the former with the latter to form an ensemble prediction has a slightly better performance across all $K$-shot adaptation cases compared to using each backbone individually. Ensemble models are well-known to yield more accurate predictions than individual model predictions \cite{hinton2015distilling} since individual models may be prone to bias/variance errors.

\paragraph{Performance with SwAV self-supervised task} We conduct an experiment to analyse the performance of our support set selection, however, using a different contrastive learning self-supervised task, namely, SwAV \cite{caron2020unsupervised}. SwAV is a classification self-supervised task that enforces consistent cluster assignment prediction for each data sample under different augmentations. This experiment aims to demonstrate that our support set selection is agnostic to the selected self-supervision objective. To this end, we conduct an experiment using PACS dataset, where we train the source backbone $f$ using the target domain data on the task of SwAV following the implementation of \cite{caron2020unsupervised}. Like BYOL, we use the backbone model $f^\prime$ and the projector network $q$ from SwAV in the remaining steps of the selection pipeline. Tab. \ref{tab:ab3} shows that SwAV yields even improved performance in the few-shot adaptation. The results show improvement at $K$=1-shot compared to BYOL and on-par performance at 5 and 10-shots. These results imply that our selection mechanism does solely depend on BYOL and can still function well using other self-supervision methods.

\subsection{Discussion}
Our \textit{SelectNAdapt} algorithm yields effective few-shot adaptation results compared to other selection baseline in the context of image recognition. However, as a part of the future work, support set selection for few-shot adaptation tasks such as image segmentation could be a investigated.
\section{Conclusion}
We presented a support set selection approach from the target domain data for few-shot domain adaptation. Our approach by leveraging self-supervision, pseudo-labelling, per-class clustering and the Euclidean distance as a scoring metric has effectively boosted the adaptation performance and dominated random selection as well as loss-based selection approaches, namely, entropy and MC-dropout. Furthermore, our selection approach avoids the need to access ground-truth of target data making it more practical compared to prior work. We have also compared to few-shot transfer learning baselines where again, our selection has demonstrated that proper selection of support samples is sufficient to improve the adaptation performance. We observed on three image recognition benchmarks that careful selection of the support set from the target domain data significantly impacts on few-shot domain adaptation.

\section*{Acknowledgment}
G.C. was supported by Australian Research Council through grant FT190100525.
{\small
\bibliographystyle{ieee_fullname}
\bibliography{main}
}
\clearpage

\beginsupplement
\section*{Supplementary Material} 
In the supplementary material, we show the per domain results of the PACS datasets for the support set selection methods, namely, random, entropy, MC-dropout, and our \textit{SelectNAdapt} algorithm. Results are shown in Tab. \ref{tab:supp_pvc}. We observe an overall consistently better results using our selection method compared to the baselines.

\begin{table}[ht]
\centering
\begin{tabular}{|l|l|cccc|}
\hline
\multirow{2}{*}{Method} & \multirow{2}{*}{K-shot} & \multicolumn{4}{c|}{PACS}                                                                      \\ \cline{3-6} 
                                              &                         & \multicolumn{1}{c|}{A}  & \multicolumn{1}{l|}{C} & \multicolumn{1}{c|}{P} & S \\ \hline
\multirow{3}{*}{Random}     & 1 & \multicolumn{1}{c|}{77.9} & \multicolumn{1}{c|}{80.0} & \multicolumn{1}{c|}{95.9} & 72.5 \\ \cline{2-6} 
                                              & 5                       & \multicolumn{1}{c|}{85.0} & \multicolumn{1}{c|}{83.3}    & \multicolumn{1}{c|}{96.5}  & 81.5   \\ \cline{2-6} 
                                              & 10                      & \multicolumn{1}{c|}{86.8} & \multicolumn{1}{c|}{86.4}    & \multicolumn{1}{c|}{97.7}  & 79.5   \\ \hline
\multirow{3}{*}{Entropy}    & 1 & \multicolumn{1}{c|}{77.6} & \multicolumn{1}{c|}{78.6} & \multicolumn{1}{c|}{95.6} & 68.4 \\ \cline{2-6} 
                                              & 5                       & \multicolumn{1}{c|}{85.7} & \multicolumn{1}{c|}{84.4}    & \multicolumn{1}{c|}{97.3}  & 77.6   \\ \cline{2-6} 
                                              & 10                      & \multicolumn{1}{c|}{87.3} & \multicolumn{1}{c|}{85.8}    & \multicolumn{1}{c|}{97.4}  & 79.2   \\ \hline
\multirow{3}{*}{MC-dropout} & 1 & \multicolumn{1}{c|}{78.3} & \multicolumn{1}{c|}{79.2} & \multicolumn{1}{c|}{95.2} & 68.4 \\ \cline{2-6} 
                                              & 5                       & \multicolumn{1}{c|}{85.9} & \multicolumn{1}{c|}{83.9}    & \multicolumn{1}{c|}{97.2}  & 77.5   \\ \cline{2-6} 
                                              & 10                      & \multicolumn{1}{c|}{86.9} & \multicolumn{1}{c|}{85.8}    & \multicolumn{1}{c|}{97.2}  & 79.2   \\ \hline
\multirow{3}{*}{\textit{\textbf{Ours}}}       & 1 & \multicolumn{1}{c|}{84.6} & \multicolumn{1}{c|}{83.0} & \multicolumn{1}{c|}{97.4} & 72.9 \\ \cline{2-6} 
                                              & 5                       & \multicolumn{1}{c|}{88.6} & \multicolumn{1}{c|}{84.3}    & \multicolumn{1}{c|}{97.8}  & 81.0   \\ \cline{2-6} 
                                              & 10                      & \multicolumn{1}{c|}{89.0} & \multicolumn{1}{c|}{85.5}    & \multicolumn{1}{c|}{97.8}  & 79.4   \\ \hline
\end{tabular}
\caption{A comparison of few-shot adaptation test results using random, entropy, MC-dropout and \textit{SelectNAdapt} algorithm (\textbf{\textit{Ours}}) for the domains of PACS dataset i.e. Photo (P), Sketch (S), Art (A), Cartoon (C).}
\label{tab:supp_pvc}
\end{table}
\end{document}